\documentclass[runningheads]{llncs}
\usepackage{graphicx}
\usepackage{multirow}
\usepackage{multicol}

\usepackage{tikz}
\usepackage{comment}
\usepackage{amsmath,amssymb} % define this before the line numbering.
\usepackage{color}
\usepackage[accsupp]{axessibility}  % Improves PDF readability for those with disabilities.

\begin{document}
% \linenumbers
\pagestyle{headings}
\mainmatter
\def\ECCVSubNumber{6}  % Insert your submission number here

\title{Detecting Driver Drowsiness as an Anomaly Using LSTM Autoencoders} % Replace with your title

\begin{comment}
\titlerunning{ECCV-22 submission ID \ECCVSubNumber} 
\authorrunning{ECCV-22 submission ID \ECCVSubNumber} 
% \author{Anonymous ECCV submission}
\institute{Paper ID \ECCVSubNumber}
\end{comment}
%******************

\titlerunning{Detecting Driver Drowsiness as an Anomaly Using LSTM Autoencoders}
\author{Gülin Tüfekci\inst{1, 2, \star} \and
Alper Kayabaşı\inst{1, 2, \star} \and
Erdem Akagündüz\inst{3} \and
İlkay Ulusoy\inst{2}}
\authorrunning{G. Tüfekci et al.}
\institute{Research Center, Aselsan Inc \and
Department of Electrical and Electronics Engineering, Middle East Technical University, Ankara, Turkey \and
Graduate School of Informatics, Middle East Technical University, Ankara, Turkey\\
\email{\{gulin.tufekci,alper.kayabasi,akaerdem,ilkay\}@metu.edu.tr}}

%******************
\maketitle
\def\thefootnote{$\star$}\footnotetext{These authors contributed equally to this work}\def\thefootnote{\arabic{footnote}}
\begin{abstract}
In this paper, an LSTM autoencoder-based architecture is utilized for drowsiness detection with ResNet-34 as feature extractor. The problem is considered as anomaly detection for a single subject; therefore, only the normal driving representations are learned and it is expected that drowsiness representations, yielding higher reconstruction losses, are to be distinguished according to the knowledge of the network. In our study, the confidence levels of normal and anomaly clips are investigated through the methodology of label assignment such that training performance of LSTM autoencoder and interpretation of anomalies encountered during testing are analyzed under varying confidence rates. Our method is experimented on NTHU-DDD and benchmarked with a state-of-the-art anomaly detection method for driver drowsiness. Results show that the proposed model achieves detection rate of 0.8740 area under curve (AUC) and is able to provide significant improvements on certain scenarios.
\keywords{Driver Drowsiness Detection, LSTM Autoencoder, Video Anomaly Detection}
\end{abstract}

\section{Introduction}

Road accidents have unrecoverable outcomes affecting the lives of many human beings. According to studies, the primary cause in road accidents is indicated as the human factor \cite{humanfactor}. Human factor is defined as inattention, cognitive distractions and improper lookout; which is the inadequate actions of the driver not paying attention to the traffic and stimuli in driving environment including the road and other vehicles, or paying attention for short periods of time. Researchers have been focusing on detecting driver drowsiness/distraction from various data sources. For example, placing sensors on the driver to gather physiological patterns \cite{brain_blood}, \cite{heart} such as brain activity, variability of heartbeat, thermal imaging, respiration pattern give clues about driver's attention; yet, may affect the driver psychologically and mislead the results. Another data source is monitoring the vehicle state such as detecting lane changes or steering wheel dynamics \cite{lanechange}, etc. One of the most preferred method is to monitor the driver's face, eyes, mouth, body, hands using a camera that may give substantial clues about driver's state \cite{blink}.

\begin{figure}[thpb]
    \centering
    \includegraphics[width = \linewidth]{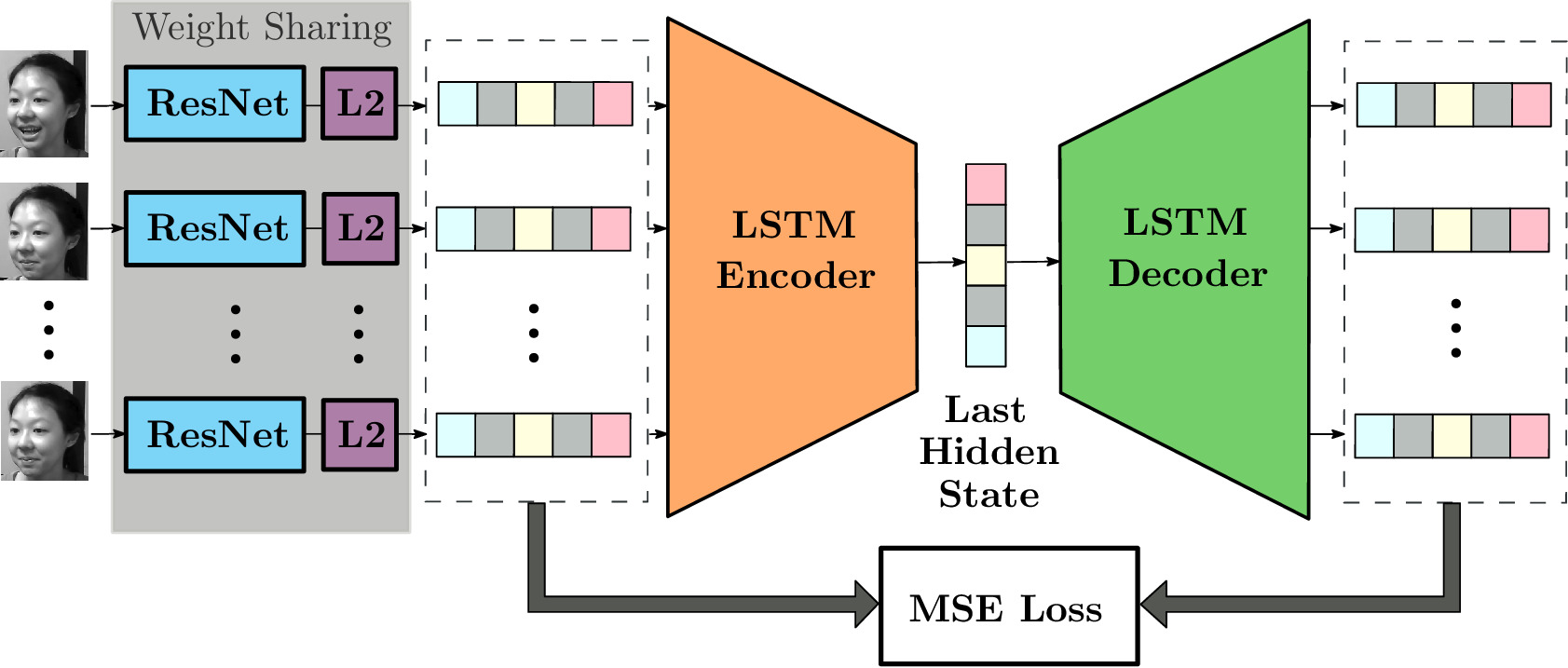}
    \caption{Architecture of the proposed method}
    \label{fig:arch}
\end{figure}

In this work, our aim is to detect driver drowsiness using a frontal camera. Detection of inattentive driver behavior is a time series problem, which is convenient for recurrent neural networks to extract hidden dependencies between time intervals. Since our approach is treating drowsy behavior as anomaly, we propose utilizing LSTM autoencoders \cite{unsupervised} as LSTMs successfully catch the long-term dependencies while making use of short-term relations. To the best of our knowledge, there are no studies that adopt LSTM autoencoders to detect driver drowsiness by treating this problem as an anomaly. Because our approach is based on anomaly detection rather than binary classification; if attentive driving is considered as normality, drowsy driving should be qualified as an anomaly. Unlike classification, where both positive and negative samples are used to train a model, only the clips that are labeled as \textit{normal} are presented to the network during training in order to learn a distribution of normal behavior and expected to produce \textit{anomaly} output whenever out-of-distribution data is present. We believe that this problem definition is more relevant to real-life scenarios, mainly because the inattentive or drowsy behavior of a driver can be mostly idiosyncratic. Collecting a training set of \textit{abnormal} behavior may never completely represent the universal set of inattentive or drowsy drivers.

In order to clarify the problem definition for drowsy and inattentive driver behavior using a frontal camera, we focus on the existing datasets and raise some important questions such as: ``\textit{how to define the abnormality degree throughout a time series data?}", or - ``\textit{if it is adequate to label a clip as ``normal" if majority of the labels belonging to the frames of the clip are normal?}" or ``\textit{how the network's performance is affected when half-confident data (having at least half of its frames labeled as normal) or full-confident data (having all of its frames labeled as normal) are used in training?}". Because the network updates itself according to the normal data that it is encountered to, the confidence level of the training data has a crucial role on specifying the knowledge of the network. There are different approaches in the literature to label the clips such as using the label of the middle frame \cite{dad} or the most frequent one \cite{literature_spatiotemporal_2}. In most studies, the labeling method is not revealed. Therefore, we believe that it is necessary to observe the effect of confidence of the clip through labeling.

In our study, the experiments are conducted on NTHU Driver Drowsiness Dataset (NTHU-DDD) \cite{nthu}. The proposed architecture, provided in Fig.~\ref{fig:arch}, contains ResNet-34 \cite{resnet} which is fine-tuned on NTHU-DDD as feature extractor and LSTM autoencoder to reconstruct the representations. The reconstruction loss is used for determining whether a clip is normal or drowsy according to a specified threshold. In order to enhance the quality of the frames, CLAHE \cite{clahe} algorithm is applied. To find a place for the proposed method in the anomaly detection literature, a state-of-the-art method is trained on NTHU-DDD as a benchmark.

The contributions of this work are twofold:
\begin{itemize}
\item LSTM autoencoders are experimented on NTHU-DDD, which is, to the best of our knowledge, the first time they are utilized for drowsiness detection using a frontal camera. State-of-the-art results are reached with learning only the \textit{normal} representations in an anomaly detection task.
\item Training strategies regarding the confidence levels of normal clips used in training are investigated and analyzed through recall and precision obtained by defining different normality and anomaly rates during test. Our experiments show that when the model is trained with low confident normal data, retrieval ability increases while reliability decreases; and vice versa for the model trained with high confident data.
\end{itemize}

\section{Related Work}

Prior to deep-learning era, hand-crafted machine learning methods were adopted to capture informative features about drowsiness such as eye blink rate \cite{blink}, eye opening \cite{eyeopening}, head orientation angle \cite{orientation}. With experience in deep neural networks, \cite{literature_spatial_1} fed the patches based on facial landmarks that are localized by MTCNN \cite{mtcnn} to distinct networks to detect drowsiness on their own dataset. Shih et al. \cite{literature_spatial_temporal_1} followed a multi stage training approach for Modified VGG-16 \cite{vgg16} and LSTM \cite{lstm} units on NTHU-DDD \cite{nthu}, taking 2nd place in the ACCV 2016 Workshop. The winner of ACCV 2016 competition \cite{literature_spatiotemporal_1} used CNN together with XGBoost \cite{xgboost} and semi supervised learning concept. \cite{literature_spatiotemporal_4} also utilized MTCNN \cite{mtcnn} to obtain facial landmarks; creating a multi-feature architecture for frames and optical flows while adopting CLAHE \cite{clahe} and Squeeze and Excitation \cite{squeeze} modules to enhance performance. 

Anomaly detection in literature can be categorized into 3 groups; which are memory-based methods \cite{mem1}, one-class classification methods \cite{one_class} and current frame reconstruction \cite{con1}. Models in the first group learn prototypes that are proxies for different normal cases, which act as basis that spans normal space so anomaly samples tend to be orthogonal to such normal space. Second group aims to find the optimal hyperplane that separates normal samples from the abnormal ones. Third group is trained to reconstruct inputs from normal cases or predicts the next normal case given previous normal cases so they are expected to be in capable of reconstructing or predicting anomaly samples. Our approach is a member of last group and anomalies occurring for single subject in a steady environment are in scope of this work. The most similar approach to our method is the work of Kopuklu et al. \cite{dad} that proposed detecting drowsiness under the concept of anomaly detection, utilizing metric learning to get the representation for normal driving and making predictions by thresholding similarity between the concerned sample and prototypes for different modalities and views on their published Driver Anomaly Dataset.

\section{The Dataset}

NTHU-DDD \cite{nthu} consists of videos belonging to 36 subjects from different ethnicities. The subjects are asked to play a driving game while pretending drowsy related actions, such as nodding, yawning, sleepiness, slow blinking, occasionally under different illumination and glasses scenarios. Drowsy frames are represented with binary labels where 1 corresponds to anomalous in our case. The videos are recorded using an infrared camera at 30 fps \cite{nthu}. Multi Task Cascaded Convolutional Neural Network \cite{mtcnn} is used for detecting faces in the dataset and cropping to eliminate redundant information. 

In order to detect drowsiness on NTHU-DDD, normal and anomaly clips should be obtained.  There are videos of 18 and 4 subjects in the training and validation sets, respectively. However, as the dataset was published for a competition, the test set is not available. The published training set of 18 subjects is divided into training and validation sets having 12 and 6 subjects respectively. The published validation set of 4 subjects is used as test set where there exists 5 videos belonging to different illumination and glasses conditions per subject. It is ensured that all sets have different subjects to prevent the network to memorize. All sets are divided into clips having 48 frames with a rate of 2 and window stride 23; so each clip is approximately 3 seconds long. By this way, around 7000 clips are obtained for the test set and around 52\% of the clips represents the anomalous case, which creates a balanced set.

\section{Methodology}

ResNet-34 \cite{resnet} which is pre-trained on ImageNet is fine-tuned on NTHU-DDD \cite{nthu}, which is a benchmark for driver drowsiness detection task, by replacing the FC layer with 2 concatenated FC layers having 128 and 2 nodes to classify the images under drowsiness label. Then, the FC layers are eliminated to obtain 512 dimensional extracted representations, which are L2 normalized to create a regularized space. Let us denote this feature extraction process as $B(\cdot) : R^{224\times 224 \times 3} \rightarrow R^{512}$ that maps $I(t)$ to $F(t)$. LSTM autoencoder reconstructs extracted features corresponding to each image, $I(t)$, in the normal clip consisting N frames, into $\hat{I}(t)$. The encoder of LSTM autoencoder, $LSTM_E$, consists of 2 layered LSTM that iteratively encodes the extracted the features to get holistic representation of spatio-temporal context of the clip at its last hidden state. Obtained context is constantly fed into $LSTM_D$, consisting 2 layered LSTM, that is responsible of decoding sequence in reverse order as \cite{unsupervised} suggests. The reconstruction loss between the input and the output of the LSTM autoencoder is computed using Mean Squared Error loss for normal clips and weights of encoder and decoder are updated to minimize the clip loss as shown in (\ref{eq:loss}) and (\ref{eq:loss2}).

\begin{equation}
\label{eq:loss}
L_{con} =  \sum_{t=1}^{N}	\: \lVert F(t) - \hat{I}(N-t+1) \rVert^2 
\end{equation}
\begin{equation}
\label{eq:loss2}
\hat{I}(t) = LSTM_D\left(LSTM_E\left(F(t)\right)\right)
\end{equation}

Since anomaly detection approach is adopted, only normal clips are used in training. To assign clip labels, normal confidence rate is defined as the controlling parameter. If the ratio of normal labeled frames over number of total frames in a clip is larger than or equal to the specified normal confidence rate, then the clip is labeled as normal. For example, with rate 1/2, at least half of the frames should be normal; generating low confident clips. With rate 1, the clip is expected to have all of its frames labeled as normal. The clips span 3 seconds of data which is sufficient to observe drowsy behaviors through such intervals. Since drowsiness clues have lingering characteristics instead of fluctuations throughout frames, the lowest normal rate value is selected as 1/2. It is reasonable to expect that the performance should increase when more confident clips are used in the training; yet, it may also mislead the network since the network learns the normal representation strictly. Hence, the performance is observed through training the model with normal rates 1/2, 2/3 and 1.

During testing, the network is subjected to anomalous clips as well as normal clips. Since the normal representations are learned throughout training, the network is expected to produce low reconstruction loss values for normal clips. However, the network is unaware of the anomaly behaviors and is expected not to successfully reconstruct the presented anomaly representations. Therefore, by applying a threshold which maximizes AUC of the corresponding receiver operating character (ROC), the test clip is predicted as normal or anomalous. In addition, anomaly confidence rate is defined so that the performance of the LSTM autoencoder can be observed under anomaly testing clips with different confidences. Anomaly rates are also tested with 1/2, 2/3 and 1.

\section{Experiment Details}

During training, frames are resized to 224$\times$224, randomly flipped horizontally, normalized by the mean and deviation for fine-tuning ResNet-34 and training the LSTM autoencoder. For fine-tuning ResNet-34 and training LSTM autoencoder; learning rates are 0.001 and 0.01, batch sizes are 128 and 4, optimizers are Adam with weight decay of 0.001 and SGD, respectively. Since NTHU-DDD consists of videos captured in night conditions, CLAHE \cite{clahe} algorithm is used for enhancing the contrast of the frames, so the pre-trained backbone ResNet-34 can extract high quality features. CLAHE is a histogram equalization technique which applies the contrast enhancement within the specified grids locally instead of using the whole image. Therefore, overexposure or underexposure caused by equalization is prevented. In the experiments, the effect is observed by applying CLAHE with different limits (5, 10) and grid sizes (8, 16), as well as without the application of CLAHE. The results regarding various normal and anomaly rates are cross compared to interpret the detection performance of the network through AUC, accuracy, recall, precision and F1.

\section{Results}

For deciding CLAHE parameters, normal and anomaly rates are set to 1/2; which is equivalent to taking the most frequent label as clip label. The AUC results are provided in Table~\ref{table:clahe}; which show that applying CLAHE is beneficial since it increases the visual perception of the network. CLAHE limit of 5 is a better contrast limit for NTHU-DDD for grid size 8. With CLAHE limit 5, the highest AUC is achieved with grid size 8.

\begin{table}[t]
\centering
\caption{AUC for different CLAHE settings on NTHU-DDD}
\label{table:clahe}
\begin{tabular}{|c|c|c|c|c|} 
\hline
CLAHE Limit & - & 10 & 5 & 5 \\ [3pt]
\hline
CLAHE Grid Size & - & 8 & 8 & 16 \\ [3pt]
\hline
AUC & 0.8240 & 0.8058 & \textbf{0.8735}  & 0.8506      \\  [3pt]
\hline
\end{tabular}
\end{table}

The test results for varying normal and anomaly confidence rates are provided in Table~\ref{table:results}. As we move towards more confident normal clips while keeping anomaly rate constant (moving towards right in rows); since representations are learnt in a relatively narrow space in a stricter sense during training, the perception ability of confident normal clips of the network increases. This effect is visible in precision; since precision defines how accurate the positive predictions are, requiring minimizing the false positives. However, when it comes to recall, the situation differs. The ability of detecting positive data (drowsy clips) of the network decreases towards right. We speculate that using low confident clips increases capacity of model to reconstruct positive examples at the expense of false positives. Therefore, in order to increase recall, using low confident clips in training is beneficial. Accuracy increases towards left; which provides the information that the network predicts more accurate but at the same time not providing any information about false positives or false negatives. Accuracy values almost remain constant for varying confidence rates; therefore, it is not proper to make a conclusion using accuracy unlike AUC, which reflects performance independent of threshold value. AUC shows that with high confident clips, false positive rate decreases while true positive rate increases. It is expected since the predictions are made for more obvious clips. As we move towards more confident anomaly clips while keeping normal rate constant (moving towards bottom in columns); the general observation is reconstruction ability of the LSTM autoencoder gets better. Since more confident anomaly clips are easier for the network to interpret, the performance tends to increase.

\begin{table*}[t]
\centering
\caption{Drowsiness detection results on NTHU-DDD}
\label{table:results}
\resizebox{\linewidth}{!}{
\begin{tabular}{|c|c|c|c|c||c|c|c||c|c|c||c|c|c||c|c|c|}
\hline
\multicolumn{2}{|c|}{} & \multicolumn{3}{c||}{AUC} & \multicolumn{3}{c||}{Accuracy (\%)}& \multicolumn{3}{c||}{Recall (\%)} & \multicolumn{3}{c||}{Precision (\%)} & \multicolumn{3}{c|}{F1 (\%)} \\ [3pt]
\cline{3-17}
\multicolumn{2}{|c|}{} & \multicolumn{3}{c||}{Normal Rates} & \multicolumn{3}{c||}{Normal Rates} & \multicolumn{3}{c||}{Normal Rates} & \multicolumn{3}{c||}{Normal Rates} & \multicolumn{3}{c|}{Normal Rates}\\ [3pt]
\cline{3-17}
\multicolumn{2}{|c|}{}  & 1/2 & 2/3 & 1  & 1/2 & 2/3 & 1  & 1/2 & 2/3 & 1  & 1/2 & 2/3 & 1& 1/2 & 2/3 & 1  \\ [3pt]
\hline
\multirow{3}{*}{Anomaly Rates} & 1/2 & 0.8740 & 0.8758 & 0.8782 & 81.58 & 81.48 & 81.44 & 79.83 & 79.56 & 78.10 & 83.39 & 84.06 & \textbf{85.41} & 81.81 & 81.75 & 81.60\\ [3pt]
\cline{2-17}
 & 2/3 & 0.8749 & 0.8768 & 0.8792 & 81.65 & 81.56 & 81.57 & 79.95 & 79.10 & 78.31 & 83.78 & 84.39 & 85.33& \textbf{81.82} & 81.66 & 81.67 \\ [3pt]
\cline{2-17}
 & 1 & 0.8763 & 0.8782 & \textbf{0.8806} & \textbf{81.79} & 81.71 & 81.78 & \textbf{80.18} & 79.34 & 78.65 & 83.51 & 84.13 & 85.10 & 81.81 & 81.66 & 81.74\\ [3pt]
\hline
\end{tabular}}
\end{table*}

\begin{figure}[t]
\centering
\begin{tabular}{cc}
    \includegraphics[width=0.12\linewidth]{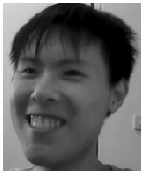} &
    \includegraphics[width=0.35\linewidth]{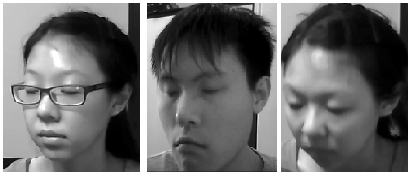} \\
    (a)&(b) \\
    \includegraphics[width=0.47\linewidth]{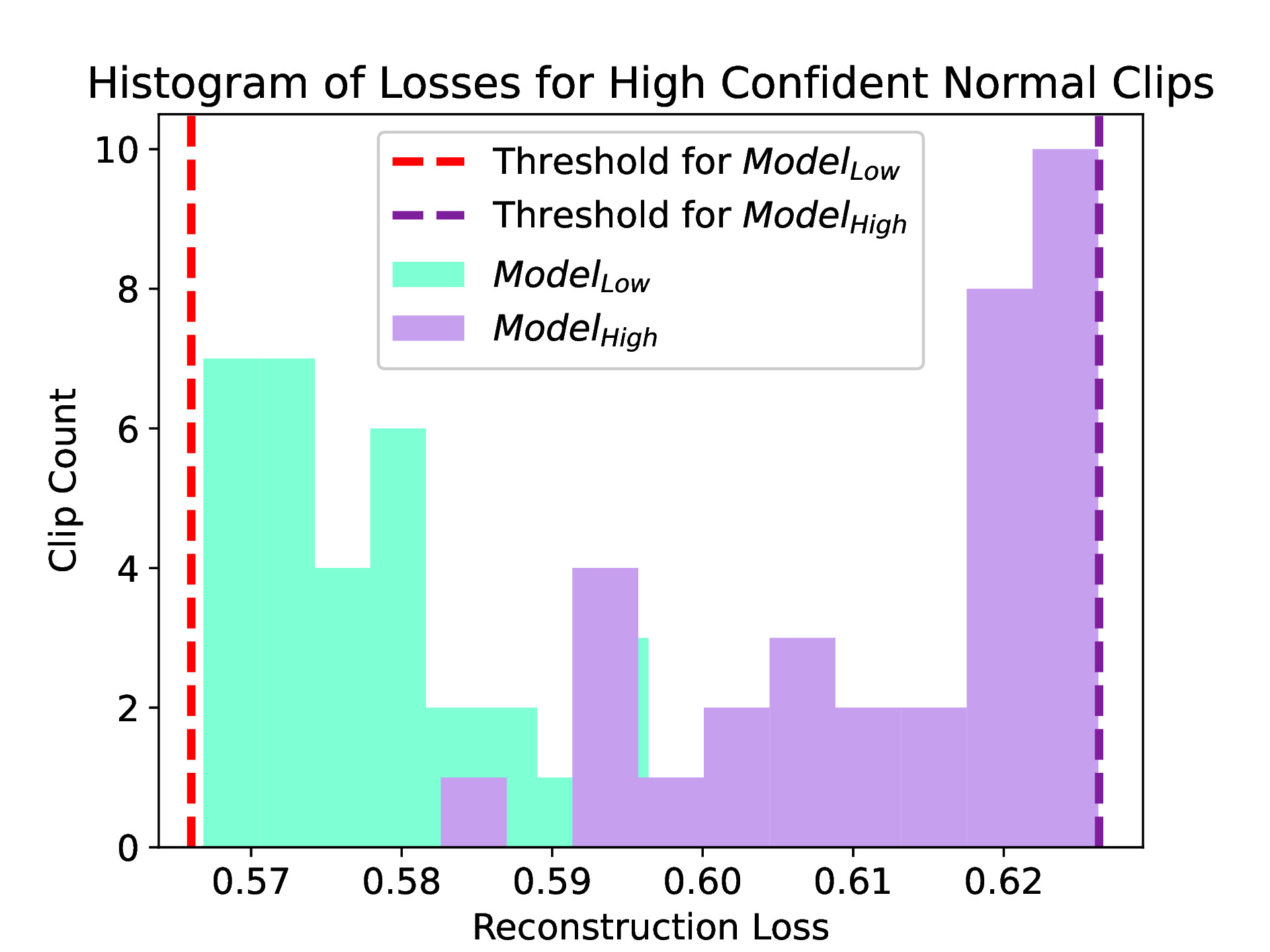} &
    \includegraphics[width=0.47\linewidth]{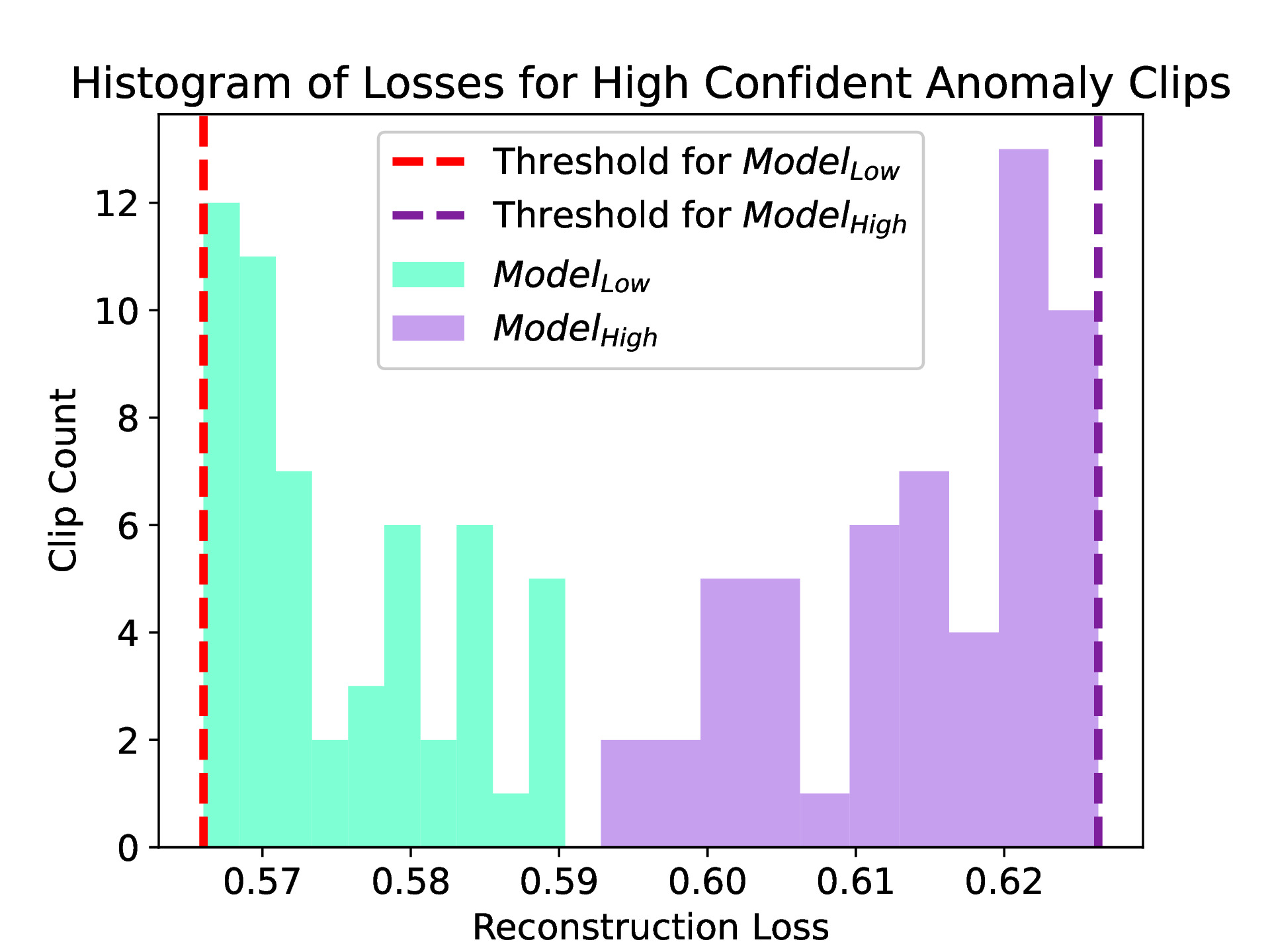} \\
    (c)&(d)
\end{tabular}
\caption{Example frames from completely normal \textbf{(a)} and completely anomaly clip \textbf{(b)} Histogram of reconstruction losses for completely normal \textbf{(c)} and completely anomalous clips \textbf{(d)}. (Best viewed in zoom)}
\label{fig:visual}
\end{figure}

\begin{figure*}[t]
    \centering
    \includegraphics[width = \linewidth]{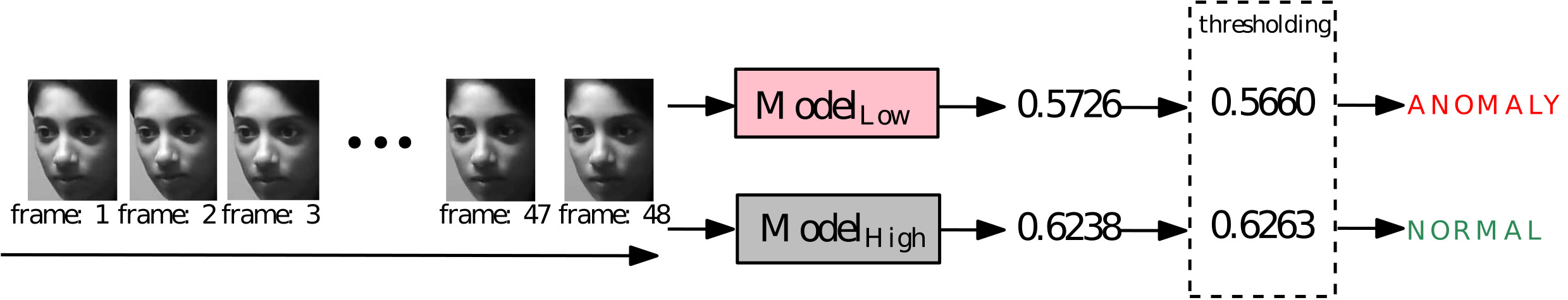}
    \caption{Evaluation of high confident normal clip by $Model_{Low}$ and $Model_{High}$}
    \label{fig:normal}
\end{figure*}

Models trained with low and high normal confidence rates are denoted by $Model_{Low}$ and $Model_{High}$ respectively. Controlled experiments are performed for analyzing the predictions of $Model_{Low}$ and $Model_{High}$ on cases where completely normal and completely anomalous clips exist. To do so, reconstruction losses are provided as histogram plots for two separate groups as shown in Fig.~\ref{fig:visual}. The first group aims to reveal the reliability of predictions for $Model_{Low}$ and $Model_{High}$ with experiments on the same completely normal clips as shown in Fig.~\ref{fig:visual} (a). As the histogram suggests in Fig.~\ref{fig:visual} (c), $Model_{High}$ predicts the clips correctly while $Model_{Low}$ mistakes the same normal clips for anomaly. This outcome is consistent with our observation that with the model trained using low confident data, false positive rate increases even when it is subjected to completely normal data. This situation is present for 33 high confident normal clips and one example is interpreted in Fig. \ref{fig:normal}. The frames of a completely normal clip is provided in the left side of the figure. $Model_{Low}$ gives an anomaly score of 0.5726 while $Model_{High}$ gives 0.6238. When these outcomes are thresholded accordingly, $Model_{Low}$ predicts the clip as anomaly (false positive case) whereas $Model_{High}$ predicts normal (true negative case). This case shows that even though $Model_{Low}$ learns the representations in a broader sense when compared to $Model_{High}$, there are still some cases where it fails to detect such high confident normal clips.

\begin{figure*}[t]
    \centering
    \includegraphics[width = \linewidth]{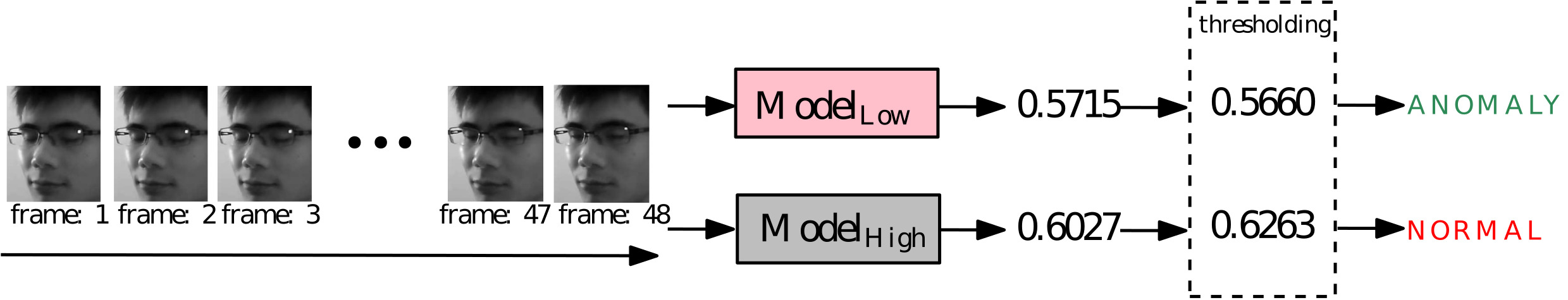}
    \caption{Evaluation of high confident anomalous clip by $Model_{Low}$ and $Model_{High}$}
    \label{fig:anomaly}
\end{figure*}

The second group, provided in Fig.~\ref{fig:visual} (d), compares the retrieval ability of $Model_{Low}$ and $Model_{High}$ on entirely anomalous clips as shown in Fig.~\ref{fig:visual} (b). In second group, $Model_{High}$ misses the samples and results in false negative predictions while $Model_{Low}$ interprets the same clips successfully. There are 55 high confident anomaly clips in the test set and one example is interpreted in Fig.~\ref{fig:anomaly}. The frames of a completely anomalous clip is provided in the left side of the figure. $Model_{Low}$ and $Model_{High}$ output scores of 0.5715 and 0.6027 respectively. When these values are thresholded accordingly, $Model_{Low}$ predicts the clip as anomaly (true positive case) whereas $Model_{High}$ predicts normal (false negative case). This case shows that although $Model_{High}$ is trained to recognize completely normal clips, it may still fail to reject such high confident anomalous clip.

Throughout varying the confidence rates, F1 scores turn out to be almost same. Hence, as a training strategy, if the aim is not to miss anomalies without the concern of increase in the false positive rate, then it is appropriate to utilize low confident clips while leaning on the recall rate. If the application area is strict about lowering the false positive rate such that only the true positives should be predicted, then precision metric is more proper to make an evaluation while using high confident clips.

\begin{table}[t]
\centering
\caption{Comparison of \cite{dad} with our method on NTHU-DDD \cite{nthu}}
\label{table:compare_nthu}
\begin{tabular}{|c|c|c|} 
\hline
Method & AUC & Accuracy \\ [3pt]
\hline

\cite{dad} Before Post-Processing & 0.8143 & 74.04   \\ [3pt]
\hline
\cite{dad} After Post-Processing  & 0.8169 & 74.32  \\ [3pt]
\hline
Ours & 0.8740 & 81.79     \\  [3pt]
\hline
\end{tabular}
\end{table}

\begin{table}[t]
\centering
\caption{Comparison of \cite{dad} with our method on DAD dataset \cite{dad}}
\label{table:compare_dad}
\begin{tabular}{|c|c|c|} 
\hline
Method & AUC & Accuracy \\ [3pt]
\hline
\cite{dad} After Post-Processing & 0.8737 & -  \\ [3pt]
\hline
Ours & 0.8434 & 74.10  \\ [3pt]
\hline
\end{tabular}
\end{table}

In our proposed method, only normal representation is learnt and drowsy clips are excluded from training data. This training strategy differs from classification. Hence, methods addressing anomaly detection are considered. For driver drowsiness detection, our method is compared with \cite{dad} due to the availability of their implementation. Their model is trained on NTHU-DDD with following changes: 32 frames with downsampling rate of 4 with front views are used to obtain clip length closed to ours and learning rate is decreased by 10-fold after 120 epochs. To make a fair comparison, our result for normal and anomaly rates of 1/2 is used. Table~\ref{table:compare_nthu} and Table~\ref{table:compare_dad} provide results for experiments on NTHU-DDD and DAD dataset respectively. For the experiments on NTHU-DDD, results of \cite{dad} before and after post-processing, which corresponds to applying running average filter on the drowsiness scores, are provided and it is shown that our method is superior than \cite{dad} on NTHU-DDD. Lastly, our method is trained on DAD dataset, yielding an AUC score of 0.8434, which is comparable with the results reported in \cite{dad}.

\section{Conclusion}

In this paper, we propose an anomaly detection-based method to detect drowsiness from camera; which consists of ResNet-34 as backbone and LSTM autoencoder that reconstructs the L2 normalized representations. The confidence levels of normal and anomalous clips are investigated to observe the effect on the knowledge of the network; which yields a training strategy depending on the application. The experiments conducted on NTHU-DDD provide an AUC value of 0.8740. As a future direction, in order to reveal the potential advantages of the proposed method, it can be applied to larger-scale data as well as multi-modal data such as depth image, data obtained from steering wheel and physiological sensors.

\bibliographystyle{splncs04}
\bibliography{egbib}
\end{document}